\documentclass[11pt,a4paper]{article}
\usepackage{emnlp2018}
\usepackage{times}
\usepackage{url}
\usepackage{latexsym}
\usepackage{graphicx}
\usepackage{multirow}
\usepackage{dsfont}
\usepackage{amsfonts}
\usepackage{colortbl}
\usepackage{xstring}

\definecolor{light-green}{RGB}{178,255,102}
\definecolor{light-red}{RGB}{255,153,153}
\definecolor{pallete-non}{RGB}{253,231,76}
\newcommand{\significant}[2]{
			\IfEqCase{#2}{%
			{+}{\cellcolor{pallete-non}#1$^{#2}$}%
			{++}{\cellcolor{green}#1$^{#2}$}%
			{-}{\cellcolor{pallete-non}#1$^{#2}$}
			{--}{\cellcolor{red}#1$^{#2}$}
    }
    [\PackageError{\significant}{Undefined option to tree: #1}{}]
    }

\renewcommand{\vec}[1]{\mathbf{#1}}

\aclfinalcopy 

\title{Transition-based Parsing with Lighter Feed-Forward Networks}

\author{
  David Vilares\\
  Universidade da Coru\~{n}a \\
  FASTPARSE Lab, LyS Group \\
  Departamento de Computaci\'{o}n \\
  Campus de Elvi\~{n}a s/n, 15071 \\ A Coru\~{n}a, Spain \\
  {\tt david.vilares@udc.es} \\
  \\\And
  Carlos G\'{o}mez-Rodr\'{i}guez \\
  Universidade da Coru\~{n}a \\
  FASTPARSE Lab, LyS Group \\
  Departamento de Computaci\'{o}n \\
  Campus de Elvi\~{n}a s/n, 15071 \\ A Coru\~{n}a, Spain \\
  {\tt carlos.gomez@udc.es}  \\}

\date{}

\begin{document}
\maketitle
\begin{abstract}

We explore whether it is possible to build lighter parsers, that are statistically equivalent to their corresponding standard version, for a wide set of languages showing different structures and morphologies. As testbed, we use the Universal Dependencies and transition-based dependency parsers trained on feed-forward networks. For these, most existing research assumes \emph{de facto standard} embedded features and relies on pre-computation tricks to obtain speed-ups. We explore how these features and their size can be reduced and whether this translates into speed-ups with a negligible impact on accuracy. The experiments show that \emph{grand-daughter} features can be removed for the majority of treebanks without a significant (negative or positive) \textsc{las} difference. They also show how the size of the embeddings can be notably reduced.

\end{abstract}

\section{Introduction}\label{intro}

 Transition-based models have achieved significant improvements in the last decade \cite{nivre2007maltparser,chen2014fast,rasooli2015yara,D17-1002}. 
Some of them already achieve a level of agreement similar to that of experts on English newswire texts \cite{berzak2016bias}, although this does not generalize to other configurations (e.g. lower-resource languages). These higher levels of accuracy often come at higher computational costs \cite{andor2016globally} and lower bandwidths, which can be a disadvantage for scenarios where speed is more relevant than accuracy \cite{GomAloVilAIRE2018}. Furthermore, running neural models on small devices for tasks such as part-of-speech tagging or word segmentation has become a matter of study \cite{pitlersmall}, showing that small feed-forward networks are suitable for these challenges. However, for parsers that are trained using neural networks, little exploration has been done beyond the application of pre-computation tricks, initially intended for fast neural machine translation \cite{devlin2014fast}, at a cost of affordable but larger memory. 

\paragraph{Contribution} We explore efficient and light dependency parsers for languages with a variety of structures and morphologies. We rely on neural feed-forward dependency parsers, since their architecture offers a competitive \emph{accuracy vs bandwidth} ratio and they are also the inspiration for more complex parsers, which also rely on embedded features but previously processed by bidirectional \textsc{lstm}s \cite{kiperwasser2016simple}. In particular, we study if the \emph{de facto standard} embedded features and their sizes can be reduced without having a significant impact on their accuracy. Building these models is of help in downstream applications of natural language processing, such as those running on small devices and also of interest for syntactic parsing itself, as it makes it possible to explore how the same configuration affects different languages.
This study is made on the Universal Dependencies v2.1, a testbed that allows us to compare a variety of languages annotated following common guidelines. This also makes it possible to extract a robust and fair comparative analysis.

\section{Related Work}

\subsection{Computational efficiency} The usefulness of dependency parsing is partially thanks to the efficiency of existing transition-based algorithms, although to the date it is an open question which algorithms suit certain languages better. To predict projective structures, a number of algorithms that run in $\mathcal{O}(n)$ with respect to the length of the input string are available. Broadly speaking, these parsers usually keep two structures: a stack (containing the words that are waiting for some arcs to be created) and a buffer (containing words awaiting to be processed). The \textsc{arc-standard} parser \cite{Nivre:2004:IDD:1613148.1613156} follows a strictly bottom-up strategy, where a word can only be assigned a head (and removed from the stack) once every daughter node has already been processed. The \textsc{arc-eager} parser avoids this restriction by including a specific transition for the reduce action. The \textsc{arc-hybrid} algorithm \cite{kuhlmann2011dynamic} mixes characteristics of both algorithms. More recent algorithms, such as \textsc{arc-swift}, have focused on the ability to manage \emph{non-local} transitions \cite{qi-manning:2017:Short} to reduce the limitations of transition-based parsers with respect to graph-based ones \cite{mcdonald2005non,dozat2016deep}, that consider a more global context. To manage non-projective structures, there are also different options available. The \newcite{covington2001fundamental} algorithm runs in $\mathcal{O}(n^2)$ in the worst case, by comparing the word in the top of the buffer with a subset of the words that have been already processed, deciding whether or not to create a link with each of them. More efficient algorithms such as \textsc{swap} \cite{nivre2009non} manage 
non-projectivity by learning when to swap pairs of words that are involved in a crossing arc, transforming it into a projective problem, with expected execution in linear time. The \textsc{2-planar} algorithm \cite{gomez2010transition} decomposes trees into at most two planar graphs, which can be used to implement a parser that runs in linear time. The \textsc{non-local covington} algorithm \cite{nonlocalcovington} combines the advantages of the wide coverage of the  \newcite{covington2001fundamental} algorithm with the non-local capabilities of the \newcite{qi-manning:2017:Short} transition system, running in quadratic time in the worst case.

\subsection{Fast dependency parsing strategies} Despite the advances in transition-based algorithms, dependency parsing still is the bottleneck for many applications. This is due to collateral issues such as the time it takes to extract features and the multiple calls to the classifier that need to be made. In traditional dependency parsing systems, such as MaltParser \cite{nivre2007maltparser}, the oracles are trained relying on machine learning algorithms, such as support vector machines, and hand-crafted \cite{huang2009bilingually,zhang2011transition} or automatically optimized sets of features \cite{ballesteros2012maltoptimizer}. The goal usually is to maximize accuracy, which often comes at a cost of bandwidth. In this sense, efforts were made in order to obtain speed-ups. Using linear classifiers might lead to faster parsers, at a cost of accuracy and larger memory usage \cite{nivre2010quick}. \newcite{BohnetFastParsing} illustrates that mapping the features into weights for a support vector machine is the major issue for the execution time and introduces a hash kernel approach to mitigate it. \newcite{volokh2013performance} made efforts on optimizing the feature extraction time for the \newcite{covington2001fundamental} algorithm, defining the concept of \emph{static features}, which can be reused through different configuration steps. The concept itself does not imply a reduction in terms of efficiency, but it is often employed in conjunction with the reduction of \emph{non-static features}, which causes a drop in accuracy. 

In more modern parsers, the oracles are trained using feed-forward networks \cite{Titov07Conll,chen2014fast,straka2015parsing} and sequential models \cite{kiperwasser2016simple}. In this sense, to obtain significant speed improvements it is common to use the pre-computation trick from \newcite{devlin2014fast}, initially intended for machine translation. Broadly speaking, they precompute the output of the hidden layer for each individual feature and each position in the input vector where they might occur, saving computation time during the test phase, with an affordable memory cost. \newcite{vacariu2017high} proposes an optimized parser and also includes a brief evaluation about reducing features that have a high cost of extraction, but the analysis is limited to English and three treebanks. However, little analysis has been made on determining if these features are relevant across a wide variety of languages that show different particularities. Our work is also in line with this line of research. In particular, we focus on feed-forward transition-based parsers, which already offer a very competitive accuracy vs bandwidth ratio. The models used in this work do not use any pre-computation trick, but it is worth pointing out that the insights of this paper could be used in conjunction with it, to obtain further bandwidth improvements.

\section{Motivation}

Transition-based dependency parsers whose oracles are trained using feed-forward neural networks have adopted as the \emph{de facto standard} set of features the one proposed by \newcite{chen2014fast} to parse the English and Chinese Penn Treebanks \cite{marcus1993building,xue2005penn}. 

We hypothesize this \emph{de facto standard} set of features and the size of the embeddings used to represent them can be reduced for a wide variety of languages, obtaining significant speed-ups at a cost of a marginal impact on their performance. To test this hypothesis, we are performing an evaluation over the Universal Dependencies v2.1 \cite{UD2.1}
a wide multilingual testbed to approximate relevant features over a wide variety of languages from different families.

\section{Methods and Materials}

This section describes the parsing algorithms (\S \ref{section-dependency-algorithms}), the architecture of the feed-forward network (\S \ref{section-network}) and the treebanks (\S \ref{section-treebank}).

\subsection{Transition-based algorithms}\label{section-dependency-algorithms}

Let $w = [w_1,w_2,...,w_{|w|}]$ be an input sentence, a \emph{dependency tree} for $w$ is an edge-labeled directed tree $T=(V,A)$ where $V = \{0,1,2,\ldots,|w|\}$ is the set of nodes and $A = V \times D \times V$ is the set of labeled arcs.  Each arc $a \in A$, of the form $(i,d,j)$, corresponds to a syntactic \emph{dependency} between the words $w_i$ and $w_j$; where $i$ is the index of the \emph{head} word, $j$ is the index of the \emph{child} word and $d$ is the \emph{dependency type} representing the kind of syntactic relation between them. Each transition configuration is represented as a 3-tuple $c=(\sigma,\beta,A)$ where:

\begin{itemize}
\item $\sigma$ is a stack that contains the words that are awaiting for remaining arcs to be created. In $\sigma|i$, $i$ represents the first word of the stack.
\item $\beta$ is a buffer structure containing the words that still have not been processed (awaiting to be moved to $\sigma$. In $i|\beta$, $i$ denotes the first word of the buffer.
\item $A$ is the set of arcs that have been created.
\end{itemize}

We rely on two transition-based algorithms: the stack-based \textsc{arc-standard} \cite{nivre2008algorithms} algorithm for projective parsing and its corresponding version with the \textsc{swap} operation \cite{nivre2009non} to manage non-projective structures. The election of the algorithms is based on their computational complexity as both run in $\mathcal{O}(n)$ empirically. The set of transitions is shown in Table \ref{table-transitions}. Let $c_i=([0],\beta,\{\})$ be an initial configuration, the parser will apply transitions until a final configuration $c_f=([0],[],A)$ is reached.

\begin{table}[bpth]
\tabcolsep=0.05cm
\centering
\small{
\begin{tabular}{llll}
&\bf Transition & \bf Step t & \bf Step \emph{t+1} \\ 
\textsc{standard}&\textsc{left-arc}$_l$ & $(\sigma|i|j,\beta,A)$ & $(\sigma|j,\beta,A \cup (j,l,i))$ \\
(projective)&\textsc{right-arc}$_l$ & $(\sigma|i|j,\beta,A)$ & $(\sigma|i,\beta,A \cup (i,l,j))$ \\
&\textsc{shift} & $(\sigma,i|\beta,A)$ &  $(\sigma|i,\beta,A)$\\
\hline
\textsc{swap}&\textsc{swap} & $(\sigma|i|j,\beta,A)$&$(\sigma|j,i|\beta,A)$\\
(above +)&&&\\
\end{tabular}}
\caption{\label{table-transitions} Transitions for the projective version of the stack-based \textsc{arc-standard} algorithm and its non-projective version including the \textsc{swap} operation}
\end{table}

\subsection{Feed-forward neural network}\label{section-network}

We reproduce the \newcite{chen2014fast} architecture and more in particular the \newcite{straka2015parsing} version. These two parsers report the fastest architectures for transition-based dependency parsing (using the pre-computation trick from \newcite{devlin2014fast}), and obtain results close to the state of the art.  Let \textsc{mlp}$_\theta(\vec{v})$ be an abstraction of our multilayered perceptron parametrized by $\theta$, the output for an input $\vec{v}$ (in this paper, a concatenation of embeddings, as described in \S \ref{section-experiments}) is computed as:
\begin{equation}
\textsc{mlp}_\theta(\vec{v}) = \mathit{softmax}(\vec{W_2} \cdot \mathit{relu}(\vec{W_1} \cdot \vec{v} + \vec{b_1}) + \vec{b_2})
\end{equation}
where $\vec{W_i}$ and $\vec{b_i}$ are the weights and bias tensors to be learned at the $i$th layer and $\mathit{softmax}$ and $\mathit{relu}$  correspond to the activation functions in their standard form.

\subsection{Universal Dependencies v2.1}\label{section-treebank}

Universal dependencies (UD) v2.1 \cite{UD2.1} is a set of 101 dependency treebanks for up to 60 different languages. They are labeled in the CoNLLU format, heavily inspired in the CoNLL format \cite{buchholz2006conll}. For each word in a sentence there is available the following information: \textsc{id}, \textsc{word}, \textsc{lemma}, \textsc{upostag} (universal postag, available for all languages), \textsc{xpostag} (language-specific postag, available for some languages), \textsc{feats} (additional morphosyntactic information, available for some languages), \textsc{head}, \textsc{deprel} and other optional columns with additional information. 

In this paper, we are only considering experiments on the \emph{unsuffixed} treebanks (where UD\_English is an unsuffixed treebank and UD\_English-PUD is a suffixed treebank). The motivation owes to practical issues and legibility of tables and discussions.

\section{Experiments}\label{section-experiments}

We followed the training configuration proposed by \newcite{straka2015parsing}.
All models where trained using mini-batches (size=10) and stochastic gradient descent (\textsc{sgd}) with exponential decay ($lr=0.02$, decay computed as  $lr \times e^{-0.2 \times epoch}$). Dropout was set to 50\%. With our implementation dropout was observed to work better than regularization with less effort in terms of tuning. We used internal embeddings, initialized according to a Glorot uniform \cite{glorot2010understanding}, which are learned together with the oracle during the training phase. In the experiments we use no external embeddings, following the same criteria as \newcite{straka2015parsing}. The aim was to evaluate all parsers under a homogeneous configuration, and high-quality external embeddings may be difficult to obtain for some languages. 

The experiments explore two paths: (1) is it possible to reduce the number of features without a significant loss in terms of accuracy? and (2) is it possible to reduce the size of the embeddings representing those features, also without causing significant loss in terms of accuracy? To evaluate this, we used as baseline the following configuration.

\subsection{Baseline configuration}

This configuration reproduces that of \newcite{straka2015parsing} which is basically a version of the \newcite{chen2014fast} parser whose features were specifically adapted to the \textsc{ud} treebanks:

\paragraph{De facto standard features} The initial set of features, which we call the \emph{de facto standard} features, is composed of: \textsc{form}, \textsc{upostag} and \textsc{feats} for the first 3 words in $\beta$ and the first 3 words of $\sigma$. The \textsc{form}, \textsc{upostag}, \textsc{feats} and \textsc{deprel}\footnote{Once it has been assigned} for the 2 leftmost and rightmost children of the first 2 words in $\sigma$. And the  \textsc{form}, \textsc{upostag}, \textsc{feats} and \textsc{deprel} of the leftmost of the leftmost and rightmost of the rightmost children of the first 2 words in $\sigma$. This makes a total of 18 elements and 66 different features.  In the case of UD treebanks, it is worth noting that for some languages the \textsc{feats} features are not available. We thought of two strategies in this situation: (1) not to consider any \textsc{feats} vector as input or (2) assume that a dummy input vector is given to represent the \textsc{feats} of an element of the tree. The former would be more realistic in a real environment, but we believe the latter offers a fairer comparison of speeds and memory costs, as the input vector is homogeneous across all languages. Thus, this is the option we have implemented. The dummy vector is expected to be given no relevance by the neural network during the training phase. We also rely on gold \textsc{upostag}s and \textsc{feats} to measure the impact of the reduced features and their reduced size in an isolated environment.\footnote{The use of predicted PoS-tags and/or tokenization would make harder to measure which is the actual impact of using different features and size of embeddings.}

\paragraph{Size of the embeddings}  The embedding size for the \textsc{form} features is set to 50 and for the \textsc{upostag}, \textsc{feats} and \textsc{deprel} features it is set to 20. Given an input configuration, the final dimension of the input vector is 1860: 540 dimensions from directly accessible nodes in $\sigma$ and $\beta$, 880 dimensions corresponding to daughter nodes and 440 dimensions corresponding to grand-daughter nodes.

\paragraph{Metrics} We use \textsc{las} (Labeled Attachment Score) to measure the performance. To determine whether the gain or loss with respect to the \emph{de facto standard} features is significant or not, we used Bikel's randomized parsing evaluation comparator ($p < 0.05$), a  stratified shuffling significance test. The null hypothesis is that the two outputs are produced by equivalent models and so the scores are equally likely.  To refute it, it first measures the difference obtained for a metric by the two models. Then, it shuffles scores of individual sentences between the two models and recomputes the evaluation metrics, measuring if the new difference is 
smaller
than the original one, which is an indicator that the outputs are significantly different. \emph{Thousands of tokens parsed per second} is the metric used to compare the speed between different feature sets. To diminish the impact of running time outliers, this is averaged across five runs. 

\paragraph{Hardware} All models were run on the test set on a single thread on a Intel(R) Core(TM) i7-7700 CPU @ 3.60GHz. 

\paragraph{No precomputation trick} All the parsers proposed in this work do not use the precomputation trick from \newcite{devlin2014fast}. There is no major reason for this, beyond measuring the impact of the strategies in a simple scenario. We would like to remark that the speed-ups obtained here by reducing the number of features could also be applied to the parsers implementing this precomputation trick, in the sense that the feature extraction time is lower. No time will be further gained in terms of computation of the hidden activation values. However, in this context, at least in the case of the \newcite{chen2014fast} parser, the pre-computation trick is only applied to the 10\,000 most common words. The experiments here proposed are also useful to save memory resources, even if the trick is used.

\subsection{Reducing the number of features}\label{section-reduced-features}

Table \ref{reduced-features} shows the impact of ignoring features that have a larger cost of extraction, i.e., daughter and grand-daughter nodes, for both the \textsc{arc-standard} and \textsc{swap} algorithms. It compares three sets of features in terms of performance and speed: (1) \emph{de facto standard} features, (2) \emph{no grand-daughter} (\textsc{no-gd}) features (excluding every leftmost of leftmost and rightmost of rightmost feature) and (3) \emph{no daughter} (\textsc{no-gd/d}) features (excluding every daughter and grand-daughter feature from nodes of $\sigma$).

\paragraph{Impact of using the \textsc{no-gd} feature set} The results show that these features can be removed without causing a significant difference in most of the cases. In the case of the \textsc{arc-standard} algorithm, for 
48
out of 52 treebanks there is no significant accuracy loss with respect to the \emph{de facto standard} features. In fact, for 22 treebanks there was a gain with respect to the original set of features, from which 5 of them were statistically significant. With respect to \textsc{swap}, we observe similar tendencies. For 
39
out of 52 treebanks there is no loss (or the loss is again not statistically significant). There is however a larger number of differences that are statistically significant, both gains (11) and losses (13). On average, the \textsc{arc-standard} models trained with these features lost 0.1 \textsc{las} points with respect to the original models, while the average speed-up was $\sim$23\%. The models trained with \textsc{swap} gained instead 0.15 points and the bandwidth increased by $\sim$28\%.

\paragraph{Impact of the \textsc{no-gd/d} features} As expected, the results show that removing \emph{daughter} features in conjunction with \emph{grand-daughter} causes a big drop in performance for the vast majority of cases (most of them statistically significant). Due to this issue and despite the (also expected) larger speed-ups, we are not considering this set of features for the next section.

\begin{table*}[hbpt]
                \centering
                \small{
                \begin{tabular}{l|cc|cc|cc||cc|cc|cc|ccc}
                \tabcolsep=0.01cm
                
                &\multicolumn{6}{c|}{\bf \textsc{arc-standard}}&\multicolumn{6}{c|}{\bf \textsc{swap}}\\
                \hline
\bf \multirow{2}{*}{Treebank}&\multicolumn{2}{c|}{\bf \textsc{standard}}&\multicolumn{2}{c|}{\bf \textsc{no-gd/d}}&\multicolumn{2}{c||}{\bf \textsc{no-gd}}&\multicolumn{2}{c|}{\bf \textsc{standard}}&\multicolumn{2}{c|}{\bf \textsc{no-gd/d}}&\multicolumn{2}{c|}{\bf \textsc{no-gd}}\\

&\bf LAS&\bf kt/s&\bf LAS&\bf kt/s&\bf LAS&\bf kt/s&\bf LAS&\bf kt/s&\bf LAS&\bf kt/s&\bf LAS&\bf kt/s\\
\hline
Afrikaans&82.72&3.3&\significant{71.67}{--}&8.4&\significant{82.42}{-}&4.0&82.55&3.0&\significant{70.59}{--}&7.7&\significant{82.96}{+}&3.8\\ 
Anc Greek&56.85&3.5&\significant{50.27}{--}&8.9&\significant{56.63}{-}&4.3&58.97&2.9&\significant{51.36}{--}&7.8&\significant{58.48}{-}&3.8\\ 
Arabic&77.46&3.1&\significant{70.69}{--}&7.7&\significant{77.87}{+}&3.7&76.77&3.0&\significant{70.4}{--}&7.4&\significant{77.5}{++}&3.7\\ 
Basque&74.26&3.6&\significant{68.13}{--}&9.0&\significant{74.05}{-}&4.4&73.98&3.2&\significant{67.31}{--}&8.4&\significant{72.44}{--}&3.8\\ 
Belarusian&70.12&2.4&\significant{61.43}{--}&5.8&\significant{67.73}{-}&2.9&69.75&2.4&\significant{62.81}{--}&5.6&\significant{70.33}{+}&2.9\\ 
Bulgarian&88.42&3.4&\significant{77.88}{--}&8.4&\significant{88.24}{-}&4.1&87.95&3.2&\significant{77.41}{--}&8.2&\significant{87.98}{+}&4.2\\ 
Catalan&87.57&3.4&\significant{76.79}{--}&8.9&\significant{87.5}{-}&4.2&87.01&3.1&\significant{76.48}{--}&8.4&\significant{87.06}{+}&3.9\\ 
Chinese&79.23&3.2&\significant{64.66}{--}&8.3&\significant{79.2}{-}&4.0&78.26&3.2&\significant{62.74}{--}&8.0&\significant{78.8}{+}&4.0\\ 
Coptic&78.68&1.9&\significant{71.32}{--}&4.9&\significant{76.0}{--}&2.3&77.25&1.3&\significant{70.08}{--}&3.1&\significant{77.44}{+}&1.5\\ 
Croatian&81.23&3.2&\significant{72.11}{--}&7.8&\significant{81.4}{+}&3.8&80.63&3.0&\significant{70.54}{--}&7.6&\significant{81.39}{++}&3.6\\ 
Czech&85.74&3.5&\significant{78.1}{--}&8.3&\significant{86.09}{++}&4.2&85.55&3.4&\significant{77.9}{--}&7.9&\significant{85.42}{-}&4.2\\ 
Danish&80.93&3.1&\significant{70.54}{--}&7.3&\significant{81.28}{+}&3.7&79.79&2.9&\significant{65.4}{--}&7.1&\significant{79.55}{-}&3.6\\ 
Dutch&78.67&3.3&\significant{66.82}{--}&8.4&\significant{79.41}{+}&4.1&77.02&3.1&\significant{64.83}{--}&7.8&\significant{78.15}{++}&3.8\\ 
English&84.16&3.6&\significant{72.68}{--}&8.8&\significant{84.42}{+}&4.4&83.19&3.6&\significant{72.76}{--}&8.7&\significant{84.09}{++}&4.4\\ 
Estonian&81.57&3.1&\significant{72.63}{--}&7.6&\significant{81.74}{+}&3.8&80.65&2.9&\significant{72.68}{--}&6.6&\significant{81.06}{+}&3.7\\ 
Finnish&81.25&3.3&\significant{69.08}{--}&8.0&\significant{82.08}{++}&4.1&81.47&3.3&\significant{69.36}{--}&7.8&\significant{80.4}{--}&3.9\\ 
French&84.65&3.0&\significant{73.15}{--}&7.2&\significant{84.83}{+}&3.5&83.54&2.7&\significant{72.31}{--}&6.8&\significant{83.27}{-}&3.5\\ 
Galician&80.51&3.5&\significant{68.82}{--}&8.5&\significant{80.29}{-}&4.2&79.85&3.3&\significant{69.25}{--}&8.3&\significant{80.01}{+}&4.2\\ 
German&79.86&3.3&\significant{72.3}{--}&8.1&\significant{79.67}{-}&4.1&78.52&3.1&\significant{70.69}{--}&8.2&\significant{77.65}{--}&3.4\\ 
Gothic&74.57&3.2&\significant{66.19}{--}&8.0&\significant{74.18}{-}&3.8&72.92&2.7&\significant{65.92}{--}&7.6&\significant{73.19}{+}&3.4\\ 
Greek&84.71&3.1&\significant{77.53}{--}&7.8&\significant{85.07}{+}&3.7&84.13&3.0&\significant{76.54}{--}&7.5&\significant{84.21}{+}&3.8\\ 
Hebrew&82.16&3.2&\significant{67.9}{--}&7.9&\significant{82.63}{+}&3.8&81.87&3.1&\significant{68.77}{--}&7.6&\significant{82.06}{+}&4.0\\ 
Hindi&90.8&3.5&\significant{81.8}{--}&8.8&\significant{90.69}{-}&4.3&90.46&3.2&\significant{80.5}{--}&8.3&\significant{90.01}{--}&4.0\\ 
Hungarian&73.34&3.1&\significant{66.39}{--}&7.0&\significant{73.14}{-}&3.5&72.33&2.9&\significant{66.88}{--}&6.8&\significant{73.36}{++}&3.1\\ 
Indonesian&79.47&2.9&\significant{62.58}{--}&7.3&\significant{79.3}{-}&3.5&78.86&2.8&\significant{63.62}{--}&7.1&\significant{79.07}{+}&3.5\\ 
Irish&60.07&2.8&\significant{52.66}{--}&7.4&\significant{59.94}{-}&3.5&61.82&2.8&\significant{54.28}{--}&7.0&\significant{60.3}{--}&3.4\\ 
Italian&89.21&2.9&\significant{78.16}{--}&7.2&\significant{89.33}{+}&3.3&88.34&2.9&\significant{78.13}{--}&7.0&\significant{88.81}{++}&3.6\\ 
Japanese&92.16&3.3&\significant{74.2}{--}&8.6&\significant{92.19}{+}&4.1&91.95&3.3&\significant{74.09}{--}&9.0&\significant{91.91}{-}&4.2\\ 
Kazakh&22.78&3.4&\significant{16.1}{--}&8.9&\significant{27.41}{++}&4.3&29.32&3.4&\significant{20.47}{--}&8.7&\significant{33.0}{++}&4.1\\ 
Korean&60.84&3.5&\significant{46.27}{--}&8.9&\significant{60.13}{-}&4.4&60.46&3.5&\significant{47.63}{--}&8.8&\significant{57.98}{--}&4.3\\ 
Latin&43.31&3.3&\significant{41.33}{--}&7.8&\significant{44.16}{++}&3.9&47.11&2.6&\significant{45.33}{--}&7.1&\significant{46.54}{-}&3.4\\ 
Latvian&75.14&3.4&\significant{64.88}{--}&8.3&\significant{75.36}{+}&4.1&74.73&3.3&\significant{65.11}{--}&8.1&\significant{75.55}{++}&4.1\\ 
Lithuanian&42.74&1.5&\significant{40.75}{-}&3.5&\significant{42.64}{-}&1.8&46.79&1.4&\significant{38.21}{--}&3.4&\significant{43.4}{--}&1.8\\ 
Marathi&66.02&1.7&\significant{61.89}{--}&4.4&\significant{62.14}{--}&2.1&65.05&1.7&\significant{59.95}{--}&4.2&\significant{66.26}{+}&2.2\\ 

Old Church&78.33&3.3&\significant{71.07}{--}&8.4&\significant{78.97}{++}&4.1&79.48&3.0&\significant{69.65}{--}&8.3&\significant{79.76}{+}&3.8\\ 

Slavonic&&&\cellcolor{red}&&\cellcolor{green}&&&&\cellcolor{red}&&\cellcolor{pallete-non}&\\ 

Persian&82.1&3.1&\significant{66.16}{--}&7.6&\significant{81.1}{--}&3.8&80.79&3.0&\significant{65.35}{--}&7.3&\significant{81.08}{+}&3.8\\ 
Polish&90.92&3.5&\significant{83.03}{--}&8.8&\significant{90.9}{-}&4.3&90.49&3.5&\significant{83.46}{--}&8.7&\significant{90.29}{-}&4.4\\ 
Portuguese&86.27&3.1&\significant{74.09}{--}&8.3&\significant{86.58}{+}&4.0&83.87&2.7&\significant{71.84}{--}&7.0&\significant{85.23}{++}&3.6\\ 
Romanian&82.12&3.3&\significant{68.89}{--}&8.2&\significant{81.73}{-}&4.1&80.92&3.3&\significant{68.82}{--}&7.9&\significant{81.05}{+}&4.1\\ 
Russian&79.47&3.0&\significant{70.0}{--}&7.5&\significant{79.14}{-}&3.7&78.55&2.9&\significant{68.63}{--}&7.4&\significant{77.62}{--}&3.7\\ 
Serbian&84.9&3.3&\significant{76.57}{--}&8.4&\significant{85.17}{+}&4.0&85.8&3.2&\significant{76.04}{--}&7.7&\significant{85.64}{-}&4.0\\ 
Slovak&85.54&3.2&\significant{76.44}{--}&7.8&\significant{85.45}{-}&4.0&84.96&3.2&\significant{77.72}{--}&8.2&\significant{84.25}{--}&3.9\\ 
Slovenian&88.73&3.2&\significant{78.06}{--}&7.3&\significant{88.74}{+}&3.8&89.35&3.0&\significant{77.67}{--}&7.1&\significant{88.31}{--}&3.7\\ 
Spanish&85.16&2.8&\significant{71.78}{--}&6.7&\significant{83.75}{--}&3.5&84.32&2.7&\significant{71.78}{--}&7.2&\significant{82.96}{--}&3.5\\ 
Swedish&83.73&3.4&\significant{71.21}{--}&8.6&\significant{83.63}{-}&4.2&84.37&3.4&\significant{69.96}{--}&8.5&\significant{83.78}{--}&4.3\\ 
(Sw) Sign&23.4&1.1&\significant{25.53}{+}&2.5&\significant{22.7}{-}&1.3&10.64&0.9&\significant{23.05}{++}&2.4&\significant{21.99}{++}&1.2\\ 

Language&&&\cellcolor{pallete-non}&&\cellcolor{pallete-non}&&&&\cellcolor{green}&&\cellcolor{green}&\\ 

Tamil&69.18&2.0&\significant{66.47}{--}&4.8&\significant{69.58}{+}&2.4&71.04&1.7&\significant{67.77}{--}&4.7&\significant{71.09}{+}&2.4\\ 
Telugu&75.17&1.5&\significant{74.76}{-}&3.1&\significant{74.48}{-}&1.6&74.2&1.4&\significant{75.45}{+}&3.0&\significant{75.03}{++}&1.6\\ 
Turkish&59.51&3.2&\significant{53.47}{--}&7.4&\significant{59.29}{-}&3.9&60.32&3.0&\significant{53.14}{--}&7.7&\significant{59.35}{-}&3.8\\ 
Ukrainian&81.29&3.2&\significant{71.95}{--}&7.9&\significant{81.71}{+}&3.8&81.8&3.0&\significant{69.82}{--}&7.6&\significant{81.19}{--}&3.9\\ 
Urdu&83.12&3.3&\significant{71.84}{--}&8.5&\significant{83.03}{-}&4.1&81.29&2.9&\significant{69.42}{--}&7.9&\significant{80.93}{-}&3.4\\ 
Vietnamese&64.73&3.4&\significant{54.71}{--}&9.0&\significant{64.39}{-}&4.3&64.35&3.5&\significant{55.51}{--}&8.8&\significant{63.71}{-}&4.3\\ 
\hline
\hline
\textsc{average}&75.67&3.0&66.42&7.5&75.57&3.7&75.29&2.8&66.07&7.2&75.44&3.6\\

\end{tabular}}
\caption{\label{reduced-features} Performance for the (1) de facto standard, (2) \textsc{no-gd/d} and (3) \textsc{no-gd} set of features, when used to train oracles with the \textsc{arc-standard} and \textsc{swap} algorithms. Red cells indicate a significant loss (-\,-) with respect to the baseline, the yellow ones a non-significant gain(+)/loss (-) and the green ones a significant gain (++).} 
\end{table*}

\begin{table*}[hbpt]
                \centering
                \small{
                \begin{tabular}{l|cc|cc|cc|cc|cc|cc|}
                \tabcolsep=0.01cm

&\multicolumn{12}{c|}{\bf \textsc{arc-standard}}\\
\hline
&\multicolumn{2}{c|}{\bf \textsc{standard}}&\multicolumn{10}{c|}{\bf \textsc{no gd}}\\
\hline
&\multicolumn{2}{c|}{\bf \,}&\multicolumn{2}{c|}{\bf size -10\%}&\multicolumn{2}{c|}{\bf size -20\%}&\multicolumn{2}{c|}{\bf size -30\%}&\multicolumn{2}{c|}{\bf size -40\%}&\multicolumn{2}{c|}{\bf size -50\%}\\
\bf Treebank&\bf LAS&\bf kt/s&\bf LAS&\bf kt/s&\bf LAS&\bf kt/s&\bf LAS&\bf kt/s&\bf LAS&\bf kt/s&\bf LAS&\bf kt/s\\
\hline

Afrikaans&82.72&3.3&\significant{82.66}{-}&4.1&\significant{82.64}{-}&4.4&82.72&4.3&\significant{82.5}{-}&4.6&\significant{83.11}{+}&4.5\\ 
Anc Greek&56.85&3.5&\significant{57.03}{+}&4.4&\significant{56.52}{-}&4.6&\significant{56.56}{-}&4.6&\significant{56.24}{--}&4.8&\significant{56.87}{+}&4.8\\ 
Arabic&77.46&3.1&\significant{77.24}{--}&3.8&\significant{76.55}{--}&4.0&\significant{77.97}{++}&4.0&\significant{77.18}{-}&4.2&\significant{77.41}{-}&4.7\\ 
Basque&74.26&3.6&\significant{74.78}{++}&4.5&\significant{74.05}{-}&4.7&\significant{74.12}{-}&4.7&\significant{73.8}{-}&4.9&\significant{74.21}{-}&4.9\\ 
Belarusian&70.12&2.4&\significant{68.67}{-}&2.9&\significant{68.74}{-}&3.0&\significant{68.02}{--}&3.0&\significant{69.18}{-}&3.1&\significant{66.86}{--}&3.3\\ 
Bulgarian&88.42&3.4&\significant{87.62}{--}&4.2&\significant{87.95}{-}&4.5&\significant{87.58}{--}&4.4&\significant{87.9}{--}&4.6&\significant{87.53}{--}&4.6\\ 
Catalan&87.57&3.4&\significant{86.77}{--}&4.3&\significant{87.63}{+}&4.7&\significant{87.22}{--}&4.6&\significant{87.28}{--}&4.8&\significant{87.35}{--}&4.7\\ 
Chinese&79.23&3.2&\significant{79.0}{-}&4.1&\significant{79.31}{+}&4.3&\significant{79.15}{-}&4.3&\significant{79.13}{-}&4.5&\significant{78.8}{-}&4.4\\ 
Coptic&78.68&1.9&\significant{76.58}{-}&2.4&\significant{78.68}{+}&2.5&\significant{79.73}{+}&2.5&\significant{77.25}{-}&2.6&\significant{75.62}{--}&2.5\\ 
Croatian&81.23&3.2&\significant{80.76}{-}&3.9&\significant{81.44}{+}&4.1&\significant{81.2}{-}&4.0&\significant{81.58}{+}&4.2&\significant{80.5}{--}&4.3\\ 
Czech&85.74&3.5&\significant{85.98}{++}&4.3&\significant{85.88}{++}&4.5&\significant{86.01}{++}&4.4&\significant{86.02}{++}&4.6&\significant{85.39}{--}&4.9\\ 
Danish&80.93&3.1&\significant{81.02}{+}&3.8&\significant{80.68}{-}&4.0&\significant{80.61}{--}&4.0&\significant{80.83}{-}&4.2&\significant{80.81}{-}&4.4\\ 
Dutch&78.67&3.3&\significant{78.63}{-}&4.2&\significant{78.63}{-}&4.4&\significant{78.87}{+}&4.4&\significant{78.13}{-}&4.6&\significant{79.36}{++}&4.6\\ 
English&84.16&3.6&\significant{84.09}{-}&4.5&\significant{83.91}{-}&4.7&\significant{84.49}{+}&4.7&\significant{84.35}{+}&4.9&\significant{83.78}{-}&4.5\\ 
Estonian&81.57&3.1&\significant{82.2}{+}&3.9&\significant{81.55}{-}&4.0&\significant{81.9}{+}&4.0&\significant{81.05}{-}&4.2&\significant{81.48}{-}&4.5\\ 
Finnish&81.25&3.3&\significant{81.37}{+}&4.2&\significant{81.8}{+}&4.4&\significant{81.52}{+}&4.3&\significant{81.71}{+}&4.5&\significant{81.03}{-}&4.6\\ 
French&84.65&3.0&\significant{84.88}{+}&3.6&\significant{85.18}{+}&3.8&\significant{84.74}{+}&3.8&\significant{84.51}{-}&3.9&\significant{85.19}{+}&4.3\\ 
Galician&80.51&3.5&\significant{79.67}{--}&4.3&\significant{80.24}{-}&4.5&\significant{79.88}{--}&4.4&\significant{80.36}{-}&4.7&\significant{80.59}{+}&4.8\\ 
German&79.86&3.3&\significant{79.0}{--}&4.2&\significant{79.54}{-}&4.5&\significant{79.65}{-}&4.4&\significant{79.54}{-}&4.6&\significant{79.38}{-}&4.4\\ 
Gothic&74.57&3.2&\significant{74.77}{+}&3.9&\significant{74.63}{+}&4.2&\significant{73.75}{--}&4.1&\significant{73.96}{--}&4.3&\significant{73.93}{-}&4.5\\ 
Greek&84.71&3.1&\significant{84.87}{+}&3.7&\significant{84.45}{-}&4.0&\significant{84.61}{-}&3.9&\significant{84.18}{--}&4.1&\significant{85.21}{+}&4.5\\ 
Hebrew&82.16&3.2&\significant{81.94}{-}&3.8&\significant{82.13}{-}&4.1&\significant{81.83}{-}&4.0&\significant{81.42}{-}&4.2&\significant{81.67}{-}&4.4\\ 
Hindi&90.8&3.5&\significant{90.92}{+}&4.4&\significant{90.66}{-}&4.7&\significant{90.46}{--}&4.6&\significant{90.73}{-}&4.8&\significant{90.42}{--}&4.8\\ 
Hungarian&73.34&3.1&\significant{72.78}{-}&3.5&\significant{73.02}{-}&3.7&\significant{72.73}{-}&3.7&\significant{72.6}{-}&3.9&\significant{72.85}{-}&4.2\\ 
Indonesian&79.47&2.9&\significant{78.81}{--}&3.5&\significant{79.07}{-}&3.7&\significant{79.23}{-}&3.7&\significant{79.31}{-}&3.9&\significant{79.1}{--}&4.0\\ 
Irish&60.07&2.8&\significant{59.17}{--}&3.5&\significant{59.72}{-}&3.7&\significant{57.94}{--}&3.7&\significant{58.6}{--}&3.9&\significant{57.55}{--}&3.8\\ 
Italian&89.21&2.9&\significant{89.34}{+}&3.3&\significant{89.16}{-}&3.5&\significant{88.33}{--}&3.5&\significant{89.16}{-}&3.6&\significant{89.57}{+}&3.8\\ 
Japanese&92.16&3.3&\significant{92.14}{-}&4.3&\significant{91.95}{-}&4.4&\significant{91.97}{-}&4.4&\significant{92.27}{+}&4.6&\significant{91.86}{-}&4.7\\ 
Kazakh&22.78&3.4&\significant{26.79}{++}&4.5&\significant{24.82}{++}&4.8&\significant{24.22}{++}&4.7&\significant{20.17}{--}&4.9&\significant{23.64}{++}&4.7\\ 
Korean&60.84&3.5&\significant{58.97}{--}&4.5&\significant{58.8}{--}&4.6&\significant{59.89}{--}&4.7&\significant{59.85}{--}&4.8&\significant{59.37}{--}&4.9\\ 
Latin&43.31&3.3&\significant{43.59}{++}&4.0&\significant{43.45}{++}&4.2&\significant{42.19}{--}&4.2&\significant{43.84}{+}&4.4&\significant{40.22}{--}&4.5\\ 
Latvian&75.14&3.4&\significant{75.83}{++}&4.2&\significant{75.23}{+}&4.5&\significant{75.26}{+}&4.4&\significant{74.85}{-}&4.7&\significant{75.1}{-}&4.6\\ 
Lithuanian&42.74&1.5&\significant{44.06}{+}&1.8&\significant{44.43}{+}&1.9&\significant{40.75}{-}&1.8&\significant{41.98}{-}&1.9&\significant{40.94}{-}&2.0\\ 
Marathi&66.02&1.7&\significant{64.32}{-}&2.2&\significant{65.53}{-}&2.3&\significant{64.81}{-}&2.3&\significant{62.86}{-}&2.3&\significant{63.11}{-}&2.4\\ 

Old Church&78.33&3.3&\significant{78.86}{+}&4.2&\significant{79.01}{++}&4.5&\significant{78.81}{++}&4.4&\significant{78.55}{+}&4.6&\significant{78.86}{++}&4.5\\ 

Slavonic&&&\cellcolor{pallete-non}&&\cellcolor{green}&&\cellcolor{green}&&\cellcolor{pallete-non}&&\cellcolor{green}&\\

Persian&82.1&3.1&\significant{81.95}{-}&3.9&\significant{82.23}{+}&4.1&\significant{82.3}{+}&4.1&\significant{82.12}{+}&4.2&\significant{82.63}{+}&4.3\\ 
Polish&90.92&3.5&\significant{90.87}{-}&4.4&\significant{90.34}{--}&4.7&\significant{90.65}{-}&4.7&\significant{90.44}{-}&4.9&\significant{90.02}{--}&4.7\\ 
Portuguese&86.27&3.1&\significant{86.47}{+}&4.0&\significant{86.5}{+}&4.3&\significant{86.72}{+}&4.2&\significant{86.02}{-}&4.5&\significant{86.53}{+}&4.2\\ 
Romanian&82.12&3.3&\significant{81.71}{-}&4.2&\significant{80.47}{--}&4.3&\significant{81.28}{--}&4.4&\significant{81.13}{--}&4.5&\significant{81.55}{-}&4.6\\ 
Russian&79.47&3.0&\significant{79.49}{+}&3.7&\significant{79.28}{-}&3.9&\significant{79.1}{-}&3.9&\significant{79.4}{-}&4.0&\significant{79.21}{-}&4.0\\ 
Serbian&84.9&3.3&\significant{85.15}{+}&4.1&\significant{85.81}{++}&4.4&\significant{85.16}{+}&4.3&\significant{85.02}{+}&4.5&\significant{85.71}{++}&4.4\\ 
Slovak&85.54&3.2&\significant{85.07}{--}&4.1&\significant{85.52}{-}&4.4&\significant{85.02}{-}&4.3&\significant{84.49}{--}&4.5&\significant{85.06}{--}&4.5\\ 
Slovenian&88.73&3.2&\significant{88.6}{-}&3.9&\significant{88.63}{-}&4.1&\significant{88.59}{-}&4.0&\significant{88.46}{-}&4.2&\significant{88.43}{-}&4.4\\ 
Spanish&85.16&2.8&\significant{85.1}{-}&3.6&\significant{84.58}{--}&3.8&\significant{84.63}{-}&3.7&\significant{84.47}{--}&3.9&\significant{84.93}{-}&4.1\\ 
Swedish&83.73&3.4&\significant{84.22}{++}&4.3&\significant{83.91}{+}&4.6&\significant{84.0}{+}&4.5&\significant{82.73}{--}&4.8&\significant{83.51}{-}&4.6\\ 

(Sw) Sign&23.4&1.1&\significant{24.47}{+}&1.4&\significant{27.3}{+}&1.4&\significant{24.82}{+}&1.4&\significant{24.11}{+}&1.5&\significant{22.7}{-}&1.4\\ 

Language&&&\cellcolor{pallete-non}&&\cellcolor{pallete-non}&&\cellcolor{pallete-non}&&\cellcolor{pallete-non}&&\cellcolor{pallete-non}&\\

Tamil&69.18&2.0&\significant{69.83}{+}&2.4&\significant{69.28}{+}&2.5&\significant{68.58}{-}&2.5&\significant{70.04}{+}&2.6&\significant{70.14}{++}&2.7\\ 
Telugu&75.17&1.5&\significant{75.73}{+}&1.7&\significant{75.73}{+}&1.7&\significant{74.48}{-}&1.7&\significant{73.65}{-}&1.7&\significant{74.06}{-}&1.9\\ 
Turkish&59.51&3.2&\significant{60.49}{++}&4.0&\significant{59.74}{+}&4.1&\significant{59.53}{+}&4.0&\significant{59.6}{+}&4.3&\significant{59.64}{+}&4.5\\ 
Ukrainian&81.29&3.2&\significant{82.05}{++}&3.9&\significant{81.61}{+}&4.1&\significant{81.79}{+}&4.0&\significant{82.08}{++}&4.2&\significant{81.46}{+}&4.2\\ 
Urdu&83.12&3.3&\significant{83.79}{++}&4.1&\significant{83.41}{+}&4.4&\significant{83.65}{++}&4.3&\significant{83.68}{++}&4.6&\significant{83.48}{+}&4.5\\ 
Vietnamese&64.73&3.4&\significant{63.73}{--}&4.4&\significant{63.5}{--}&4.7&\significant{64.32}{-}&4.6&\significant{64.7}{-}&4.9&\significant{63.96}{--}&4.8\\ 
\hline
\hline
\textsc{average}&75.67&3.0&75.65&3.8&75.67&3.9&75.45&3.9&75.29&4.1&75.22&4.2

\end{tabular}}
\caption{\label{reduced-embeddings} \textsc{arc-standard} baseline configuration versus different runs with the \textsc{no-gd} feature set and embedding size reduction from 10\% to 50\%. See Table \ref{reduced-features} for color scheme definition.}
\end{table*}

%
%

\begin{table*}[hbpt]
                \centering
                \scriptsize{
                \begin{tabular}{l|ll|ll|ll||l|ll|ll|ll|}

&\multicolumn{2|}{c}{\bf \textsc{standard}}&\multicolumn{4}{c||}{\bf \textsc{no-gd}}&&\multicolumn{2}{c|}{\bf \textsc{standard}}&\multicolumn{4}{c|}{\bf \textsc{no-gd}}\\
\hline
\multirow{2}{*}{\bf Treebank}&&&\multicolumn{2}{c|}{\bf size -10\%}&\multicolumn{2}{c||}{\bf size -50\%}&&&&\multicolumn{2}{c|}{\bf size -10\%}&\multicolumn{2}{c|}{\bf size -50\%}\\
 & \bf LAS &\bf kt/s & \bf LAS &\bf kt/s & \bf LAS
&\bf kt/s & \bf Treebank  & \bf LAS &\bf kt/s  & \bf LAS &\bf kt/s & \bf LAS &\bf kt/s \\
\hline
Afrikaans&82.55&3.0&\significant{83.19}{+}&3.9&\significant{80.55}{--}&4.2&
Anc Greek&58.97&2.9&\significant{59.04}{+}&3.6&\significant{60.3}{++}&3.8\\ 

Arabic&76.77&3.0&\significant{76.8}{+}&3.8&\significant{76.71}{-}&4.1&
Basque&73.98&3.2&\significant{73.74}{-}&4.2&\significant{72.48}{--}&4.3\\ 

Belarusian&69.75&2.4&\significant{69.54}{-}&2.9&\significant{68.52}{-}&3.2&
Bulgarian&87.95&3.2&\significant{87.06}{--}&4.2&\significant{87.73}{-}&4.9\\ 

Catalan&87.01&3.1&\significant{87.16}{+}&3.9&\significant{86.8}{-}&4.6&
Chinese&78.26&3.2&\significant{79.76}{++}&4.1&\significant{78.19}{-}&4.6\\ 

Coptic&77.25&1.3&\significant{78.01}{+}&1.7&\significant{76.29}{-}&1.5&
Croatian&80.63&3.0&\significant{81.28}{+}&4.0&\significant{81.37}{++}&4.1\\ 

Czech&85.55&3.4&\significant{83.49}{--}&4.3&\significant{84.97}{--}&4.7&
Danish&79.79&2.9&\significant{78.79}{--}&3.7&\significant{78.55}{--}&4.0\\ 

Dutch&77.02&3.1&\significant{77.93}{+}&3.9&\significant{77.53}{+}&4.3&
English&83.19&3.6&\significant{83.61}{+}&4.5&\significant{83.92}{++}&4.9\\ 

Estonian&80.65&2.9&\significant{80.01}{-}&3.7&\significant{80.72}{+}&4.1&
Finnish&81.47&3.3&\significant{80.98}{--}&4.1&\significant{81.25}{-}&4.4\\ 

French&83.54&2.7&\significant{82.53}{--}&3.5&\significant{83.78}{+}&3.9&
Galician&79.85&3.3&\significant{80.44}{++}&4.4&\significant{80.01}{+}&4.8\\ 

German&78.52&3.1&\significant{77.53}{--}&3.5&\significant{77.6}{--}&4.2&
Gothic&72.92&2.7&\significant{72.96}{+}&3.5&\significant{71.43}{--}&3.8\\ 

Greek&84.13&3.0&\significant{83.91}{-}&3.9&\significant{84.11}{-}&4.4&
Hebrew&81.87&3.1&\significant{82.07}{+}&4.0&\significant{82.41}{+}&4.5\\ 

Hindi&90.46&3.2&\significant{89.86}{--}&4.0&\significant{89.58}{--}&4.4&
Hungarian&72.33&2.9&\significant{73.77}{++}&3.7&\significant{72.8}{+}&3.3\\ 

Indonesian&78.86&2.8&\significant{79.0}{+}&3.6&\significant{79.19}{+}&3.9&
Irish&61.82&2.8&\significant{60.67}{--}&3.5&\significant{60.88}{--}&3.9\\ 

Italian&88.34&2.9&\significant{88.39}{+}&3.4&\significant{88.51}{+}&4.0&
Japanese&91.95&3.3&\significant{92.02}{+}&4.3&\significant{91.91}{-}&4.7\\ 

Kazakh&29.32&3.4&\significant{29.64}{++}&4.3&\significant{29.77}{+}&4.8&
Korean&60.46&3.5&\significant{59.66}{--}&4.4&\significant{58.75}{--}&5.1\\ 

Latin&47.11&2.6&\significant{45.05}{--}&3.3&\significant{44.3}{--}&3.6&
Latvian&74.73&3.3&\significant{75.05}{+}&4.2&\significant{75.05}{+}&4.8\\ 

Lithuanian&46.79&1.4&\significant{44.72}{-}&1.8&\significant{44.91}{-}&1.9&
Marathi&65.05&1.7&\significant{64.81}{-}&2.2&\significant{65.53}{+}&2.4\\ 

Old Church &79.48&3.0&\significant{80.07}{+}&3.9&\significant{77.62}{--}&4.4&
Persian&80.79&3.0&\significant{81.54}{+}&3.7&\significant{80.84}{+}&4.3\\ 

Slavonic&&&&&&&&&&&&&\\

Polish&90.49&3.5&90.49&4.4&\significant{90.17}{-}&4.9&
Portuguese&83.87&2.7&\significant{83.09}{-}&2.9&\significant{84.41}{+}&3.6\\ 

Romanian&80.92&3.3&\significant{80.08}{--}&4.1&\significant{80.1}{--}&4.5&
Russian&78.55&2.9&\significant{77.78}{--}&3.7&\significant{77.75}{--}&4.1\\ 

Serbian&85.8&3.2&\significant{85.02}{--}&4.1&\significant{85.24}{-}&4.5&
Slovak&84.96&3.2&\significant{85.25}{+}&3.9&\significant{84.27}{--}&4.4\\ 

Slovenian&89.35&3.0&\significant{88.85}{--}&3.6&\significant{88.88}{--}&4.2&
Spanish&84.32&2.7&\significant{83.84}{-}&3.6&\significant{83.28}{--}&3.7\\ 

Swedish&84.37&3.4&\significant{82.8}{--}&4.3&\significant{83.72}{--}&4.8&
(Sw) Sign &10.64&0.9&\significant{21.28}{++}&1.2&\significant{17.73}{++}&1.2\\ 

&&&&&&&Language&&&&&&\\

Tamil&71.04&1.7&\significant{70.54}{-}&3.2&\significant{70.79}{-}&2.6&
Telugu&74.2&1.4&\significant{75.17}{++}&1.7&\significant{73.79}{-}&1.7\\ 

Turkish&60.32&3.0&\significant{59.71}{-}&3.9&\significant{58.45}{--}&4.3&
Ukrainian&81.8&3.0&\significant{80.66}{--}&4.0&\significant{80.86}{--}&4.4\\ 

Urdu&81.29&2.9&\significant{81.42}{+}&3.7&\significant{82.3}{++}&4.0&
Vietnamese&64.35&3.5&\significant{64.32}{-}&4.4&\significant{64.67}{+}&5.0\\ 
\hline

\end{tabular}}
             \caption{\label{reduced-embeddings-swap}  \textsc{swap} baseline configuration versus different runs with the \textsc{no-gd} feature set and embedding size reduction by a factor of 0.1 and 0.5. The average \textsc{las}/speed for the baseline is 75.29/2.8, for the \textsc{no-gd} feature set with embedding reduction by a factor of 0.1 is 75.27/3.6, and with embedding reduction by a factor of 0.5 75.02/4.0. See Table \ref{reduced-features} for color scheme definition.}
\end{table*}

\subsection{Reducing the embedding size of the selected features}\label{section-reduced-embeddings}

We now explore whether by reducing the size of the embeddings for the \textsc{form}, \textsc{postag}, \textsc{feats} and \textsc{deprel} features the models can produce better bandwidths without suffering a lack of accuracy. We run separate experiments for the \textsc{arc-standard} and \textsc{swap} algorithms, using as the starting point the \textsc{no-gd} feature set, which had a negligible impact on accuracy, as tested in Table \ref{reduced-features}. Table \ref{reduced-embeddings} summarizes the experiments when reducing the size of each embedding from 10\% to 50\%, at a step size of 10 percentage points, for the \textsc{arc-standard.} The results include information indicating whether the difference in performance is statistically significant from that obtained by the \emph{de facto} standard set. In general terms, reducing the size of the embeddings causes a small but constant drop in the performance. However, for the vast majority of languages this drop is not statistically significant. Reducing the size of the embeddings by a factor of 0.2 was the configuration with the minimum number of significant losses (6), and reducing them by a factor of 0.5 the one with the largest (14). On average, the lightest models lost 0.45 \textsc{las} points to obtain an speed-up of $\sim$40\%. Similar tendencies were observed in the case of the non-projective algorithm, whose results reducing the size of the embeddings by a factor of 0.1 and 0.5 can be found in Table \ref{reduced-embeddings-swap}.

\subsection{Discussion}

Different deep learning frameworks to build neural networks might present differences and implementation details that might cause the speed obtained empirically to differ from theoretical expectations. 

From a theoretical point of view, both tested approaches (\S \ref{section-reduced-features}, \ref{section-reduced-embeddings}) should have a similar impact, as their use directly affects the size of the input to the neural network. The smaller the input size, the lighter and faster parsers are obtained. As a side note, with respect to the case of reducing the number of features (\S \ref{section-reduced-features}), an additional speed improvement is expected, as less features need to be collected. But broadly speaking, the speed obtained by skipping half of the features should be in line with that obtained by reducing the size of the embeddings of the original features by a factor of 0.5. 

For a practical point of view, in this work we relied on \url{keras} \cite{chollet2015keras}. With respect to the part reported in \S \ref{section-reduced-features}, the experiments went as expected. Taking as examples the results for the \textsc{arc-standard} algorithm, using no \emph{grand-daughter} features implies to diminish the dimension of the input vector from 1860 dimensions to 1420, a reduction of $\sim$23\%. The average thousands of tokens parsed per second of the \emph{de facto standard} features was 3.0 and the average obtained without \emph{grand-daughter} features was 3.7, a gain of $\sim$20\%. If we also skip \emph{daughter} features and reduce the size of the input vector by $\sim$71\%, the speed increased by a factor of 2.5. Similar tendencies were observed with respect to the \textsc{swap} algorithm. When reducing the size of the embeddings (\S \ref{section-reduced-embeddings}), the obtained speed-ups were however lower than those expected in theory. In this sense, an alternative implementation or a use of a different framework could lead to reduce these times to values closer to the theoretical expectation.

Trying other neural architectures is also of high interest, but this is left as an open question for future research. In particular, in the popular \textsc{bist}-based parsers \cite{kiperwasser2016simple,de2017raw,vilares2017non}, the input is first processed by a bidirectional \textsc{lstm} \cite{hochreiter1997long} that computes an embedding for each token, taking into account its left and right context. These embeddings are then used to extract the features for transition-based algorithms, including the head of different elements and their leftmost/rightmost children. Those features are then fed to a feed-forward network, similar to the one evaluated in this work. Thus, the results of this work might be of future interest for this type of parsers too, as the output of the \textsc{lstm} can be seen as improved and better contextualized word embeddings.

\section{Conclusion}

We explored whether it is possible to reduce the number and size of embedded features assumed as \emph{de facto standard} by feed-forward network transition-based dependency parsers. The aim was to train efficient and light parsers for a vast amount of languages showing a rich variety of structures and morphologies.

To test the hypothesis we used a multilingual testbed: the Universal Dependencies v2.1. The study considered two transition-based algorithms to train the oracles: a stack-based \textsc{arc-standard} and its non-projective version, by adding the \textsc{swap} operation. We first evaluated three sets of features, clustered according to their extraction costs: (1) the \emph{de facto standard} features that usually are fed as input to feed-forward parsers and consider \emph{daughter} and \emph{grand-daughter} features, (2) a \emph{no grand-daughter} feature set and (3) a \emph{no grand-daughter/daughter} feature set. For the majority of the treebanks we found that the feature set (2) did not cause a significant loss, both for the stack-based \textsc{arc-standard} and the \textsc{swap} algorithms. We then took that set of features and reduced the size of the embeddings used to represent each feature, up to a factor of 0.5. 
The experiments also show that for both the \textsc{arc-standard} and the \textsc{swap} algorithms these reductions did not cause, in general terms, a significant loss. As a result, we obtained a set of lighter and faster transition-based parsers that achieve a better \emph{accuracy vs bandwidth} ratio than the original ones. It was observed that these improvements were not restricted to a particular language family or specific morphology.

As future work, it would be interesting to try alternative experiments to see whether reducing the size of embeddings works the same for words as for other features. Also, the results are compatible with existent optimizations and can be used together to obtain further speed-ups. Related to this, quantized word vectors \cite{2018arXiv180305651L} can save memory and be used to outperform traditional embeddings.

\section*{Acknowledgments}

We would like to thank the anonymous reviewers for their useful suggestions and detailed comments.
This work has received funding from the European
Research Council (ERC), under the European
Union's Horizon 2020 research and innovation
programme (FASTPARSE, grant agreement No
714150), from the TELEPARES-UDC project
(FFI2014-51978-C2-2-R) and the ANSWER-ASAP project (TIN2017-85160-C2-1-R) from MINECO, and from Xunta de Galicia (ED431B 2017/01). 
We gratefully acknowledge NVIDIA Corporation for the donation of a GTX Titan X GPU.

\bibliographystyle{acl_natbib_nourl}
\bibliography{biblio}

\begin{thebibliography}{39}
\expandafter\ifx\csname natexlab\endcsname\relax\def\natexlab#1{#1}\fi

\bibitem[{Andor et~al.(2016)Andor, Alberti, Weiss, Severyn, Presta, Ganchev,
  Petrov, and Collins}]{andor2016globally}
Daniel Andor, Chris Alberti, David Weiss, Aliaksei Severyn, Alessandro Presta,
  Kuzman Ganchev, Slav Petrov, and Michael Collins. 2016.
\newblock Globally normalized transition-based neural networks.
\newblock In \emph{Proceedings of the 54th Annual Meeting of the Association
  for Computational Linguistics (Volume 1: Long Papers)}, pages 2442--2452,
  Berlin, Germany. Association for Computational Linguistics.

\bibitem[{Ballesteros and Nivre(2012)}]{ballesteros2012maltoptimizer}
Miguel Ballesteros and Joakim Nivre. 2012.
\newblock Maltoptimizer: A system for maltparser optimization.
\newblock In \emph{LREC}, pages 2757--2763.

\bibitem[{Berzak et~al.(2016)Berzak, Huang, Barbu, Korhonen, and
  Katz}]{berzak2016bias}
Yevgeni Berzak, Yan Huang, Andrei Barbu, Anna Korhonen, and Boris Katz. 2016.
\newblock Anchoring and agreement in syntactic annotations.
\newblock In \emph{Proceedings of the 2016 Conference on Empirical Methods in
  Natural Language Processing}, pages 2215--2224, Austin, Texas. Association
  for Computational Linguistics.

\bibitem[{Bohnet(2010)}]{BohnetFastParsing}
Bernd Bohnet. 2010.
\newblock Very high accuracy and fast dependency parsing is not a
  contradiction.
\newblock In \emph{Proceedings of the 23rd International Conference on
  Computational Linguistics}, COLING '10, pages 89--97, Stroudsburg, PA, USA.
  Association for Computational Linguistics.

\bibitem[{Botha et~al.(2017)Botha, Pitler, Ma, Bakalov, Salcianu, Weiss,
  McDonald, and Petrov}]{pitlersmall}
Jan~A. Botha, Emily Pitler, Ji~Ma, Anton Bakalov, Alex Salcianu, David Weiss,
  Ryan McDonald, and Slav Petrov. 2017.
\newblock Natural language processing with small feed-forward networks.
\newblock In \emph{Proceedings of the 2017 Conference on Empirical Methods in
  Natural Language Processing}, pages 2879--2885. Association for Computational
  Linguistics.

\bibitem[{Buchholz and Marsi(2006)}]{buchholz2006conll}
Sabine Buchholz and Erwin Marsi. 2006.
\newblock {CoNLL-X} shared task on multilingual dependency parsing.
\newblock In \emph{Proceedings of the Tenth Conference on Computational Natural
  Language Learning}, pages 149--164. Association for Computational
  Linguistics.

\bibitem[{Chen and Manning(2014)}]{chen2014fast}
Danqi Chen and Christopher Manning. 2014.
\newblock A fast and accurate dependency parser using neural networks.
\newblock In \emph{Proceedings of the 2014 conference on empirical methods in
  natural language processing (EMNLP)}, pages 740--750.

\bibitem[{Chollet et~al.(2015)}]{chollet2015keras}
Fran\c{c}ois Chollet et~al. 2015.
\newblock Keras.
\newblock \url{https://github.com/keras-team/keras}.

\bibitem[{Covington(2001)}]{covington2001fundamental}
Michael~A Covington. 2001.
\newblock A fundamental algorithm for dependency parsing.
\newblock In \emph{Proceedings of the 39th annual ACM southeast conference},
  pages 95--102. Citeseer.

\bibitem[{Devlin et~al.(2014)Devlin, Zbib, Huang, Lamar, Schwartz, and
  Makhoul}]{devlin2014fast}
Jacob Devlin, Rabih Zbib, Zhongqiang Huang, Thomas Lamar, Richard Schwartz, and
  John Makhoul. 2014.
\newblock Fast and robust neural network joint models for statistical machine
  translation.
\newblock In \emph{Proceedings of the 52nd Annual Meeting of the Association
  for Computational Linguistics (Volume 1: Long Papers)}, volume~1, pages
  1370--1380.

\bibitem[{Dozat and Manning(2017)}]{dozat2016deep}
Timothy Dozat and Christopher~D. Manning. 2017.
\newblock Deep biaffine attention for neural dependency parsing.
\newblock In \emph{Proceedings of the 5th International Conference on Learning
  Representations}.

\bibitem[{Fern{\'a}ndez-Gonz{\'a}lez and
  G{\'o}mez-Rodr{\'i}guez(2018)}]{nonlocalcovington}
Daniel Fern{\'a}ndez-Gonz{\'a}lez and Carlos G{\'o}mez-Rodr{\'i}guez. 2018.
\newblock Non-projective dependency parsing with non-local transitions.
\newblock In \emph{Proceedings of the 2018 Conference of the North American
  Chapter of the Association for Computational Linguistics: Human Language
  Technologies, Volume 2 (Short Papers)}, pages 693--700. Association for
  Computational Linguistics.

\bibitem[{Glorot and Bengio(2010)}]{glorot2010understanding}
Xavier Glorot and Yoshua Bengio. 2010.
\newblock Understanding the difficulty of training deep feedforward neural
  networks.
\newblock In \emph{Proceedings of the Thirteenth International Conference on
  Artificial Intelligence and Statistics}, pages 249--256.

\bibitem[{G{\'o}mez-Rodr{\'i}guez et~al.(2017)G{\'o}mez-Rodr{\'i}guez,
  Alonso-Alonso, and Vilares}]{GomAloVilAIRE2018}
Carlos G{\'o}mez-Rodr{\'i}guez, Iago Alonso-Alonso, and David Vilares. 2017.
\newblock How important is syntactic parsing accuracy? an empirical evaluation
  on rule-based sentiment analysis.
\newblock \emph{Artificial Intelligence Review}.

\bibitem[{G{\'o}mez-Rodr{\'\i}guez and Nivre(2010)}]{gomez2010transition}
Carlos G{\'o}mez-Rodr{\'\i}guez and Joakim Nivre. 2010.
\newblock A transition-based parser for 2-planar dependency structures.
\newblock In \emph{Proceedings of the 48th Annual Meeting of the Association
  for Computational Linguistics}, pages 1492--1501. Association for
  Computational Linguistics.

\bibitem[{Hochreiter and Schmidhuber(1997)}]{hochreiter1997long}
Sepp Hochreiter and J{\"u}rgen Schmidhuber. 1997.
\newblock Long short-term memory.
\newblock \emph{Neural computation}, 9(8):1735--1780.

\bibitem[{Huang et~al.(2009)Huang, Jiang, and Liu}]{huang2009bilingually}
Liang Huang, Wenbin Jiang, and Qun Liu. 2009.
\newblock Bilingually-constrained (monolingual) shift-reduce parsing.
\newblock In \emph{Proceedings of the 2009 Conference on Empirical Methods in
  Natural Language Processing: Volume 3-Volume 3}, pages 1222--1231.
  Association for Computational Linguistics.

\bibitem[{Kiperwasser and Goldberg(2016)}]{kiperwasser2016simple}
Eliyahu Kiperwasser and Yoav Goldberg. 2016.
\newblock Simple and accurate dependency parsing using bidirectional {LSTM}
  feature representations.
\newblock \emph{Transactions of the Association for Computational Linguistics},
  4:313--327.

\bibitem[{Kuhlmann et~al.(2011)Kuhlmann, G{\'o}mez-Rodr{\'\i}guez, and
  Satta}]{kuhlmann2011dynamic}
Marco Kuhlmann, Carlos G{\'o}mez-Rodr{\'\i}guez, and Giorgio Satta. 2011.
\newblock Dynamic programming algorithms for transition-based dependency
  parsers.
\newblock In \emph{Proceedings of the 49th Annual Meeting of the Association
  for Computational Linguistics: Human Language Technologies-Volume 1}, pages
  673--682. Association for Computational Linguistics.

\bibitem[{{Lam}(2018)}]{2018arXiv180305651L}
M.~{Lam}. 2018.
\newblock {Word2Bits - Quantized Word Vectors}.
\newblock \emph{ArXiv e-prints}.

\bibitem[{de~Lhoneux et~al.(2017)de~Lhoneux, Shao, Basirat, Kiperwasser,
  Stymne, Goldberg, and Nivre}]{de2017raw}
Miryam de~Lhoneux, Yan Shao, Ali Basirat, Eliyahu Kiperwasser, Sara Stymne,
  Yoav Goldberg, and Joakim Nivre. 2017.
\newblock From raw text to universal dependencies-look, no tags!
\newblock \emph{Proceedings of the CoNLL 2017 Shared Task: Multilingual Parsing
  from Raw Text to Universal Dependencies}, pages 207--217.

\bibitem[{Marcus et~al.(1993)Marcus, Marcinkiewicz, and
  Santorini}]{marcus1993building}
Mitchell~P Marcus, Mary~Ann Marcinkiewicz, and Beatrice Santorini. 1993.
\newblock Building a large annotated corpus of {E}nglish: The {P}enn treebank.
\newblock \emph{Computational linguistics}, 19(2):313--330.

\bibitem[{McDonald et~al.(2005)McDonald, Pereira, Ribarov, and
  Haji{\v{c}}}]{mcdonald2005non}
Ryan McDonald, Fernando Pereira, Kiril Ribarov, and Jan Haji{\v{c}}. 2005.
\newblock Non-projective dependency parsing using spanning tree algorithms.
\newblock In \emph{Proceedings of the conference on Human Language Technology
  and Empirical Methods in Natural Language Processing}, pages 523--530.
  Association for Computational Linguistics.

\bibitem[{Nivre(2004)}]{Nivre:2004:IDD:1613148.1613156}
Joakim Nivre. 2004.
\newblock Incrementality in deterministic dependency parsing.
\newblock In \emph{Proceedings of the Workshop on Incremental Parsing: Bringing
  Engineering and Cognition Together}, IncrementParsing '04, pages 50--57,
  Stroudsburg, PA, USA. Association for Computational Linguistics.

\bibitem[{Nivre(2008)}]{nivre2008algorithms}
Joakim Nivre. 2008.
\newblock Algorithms for deterministic incremental dependency parsing.
\newblock \emph{Computational Linguistics}, 34(4):513--553.

\bibitem[{Nivre(2009)}]{nivre2009non}
Joakim Nivre. 2009.
\newblock Non-projective dependency parsing in expected linear time.
\newblock In \emph{Proceedings of the Joint Conference of the 47th Annual
  Meeting of the ACL and the 4th International Joint Conference on Natural
  Language Processing of the AFNLP: Volume 1-Volume 1}, pages 351--359.
  Association for Computational Linguistics.

\bibitem[{Nivre and Hall(2010)}]{nivre2010quick}
Joakim Nivre and Johan Hall. 2010.
\newblock A quick guide to {M}alt{P}arser optimization.
\newblock \emph{http://maltparser. org/guides/opt/quick-opt.pdf}.

\bibitem[{Nivre et~al.(2007)Nivre, Hall, Nilsson, Chanev, Eryigit, K{\"u}bler,
  Marinov, and Marsi}]{nivre2007maltparser}
Joakim Nivre, Johan Hall, Jens Nilsson, Atanas Chanev, G{\"u}l{\c{s}}en
  Eryigit, Sandra K{\"u}bler, Svetoslav Marinov, and Erwin Marsi. 2007.
\newblock Maltparser: A language-independent system for data-driven dependency
  parsing.
\newblock \emph{Natural Language Engineering}, 13(2):95--135.

\bibitem[{Nivre et~al.(2017)}]{UD2.1}
Joakim Nivre et~al. 2017.
\newblock Universal dependencies 2.1.
\newblock {LINDAT}/{CLARIN} digital library at the Institute of Formal and
  Applied Linguistics ({{\'U}FAL}), Faculty of Mathematics and Physics, Charles
  University.

\bibitem[{Qi and Manning(2017)}]{qi-manning:2017:Short}
Peng Qi and Christopher~D. Manning. 2017.
\newblock Arc-swift: A novel transition system for dependency parsing.
\newblock In \emph{Proceedings of the 55th Annual Meeting of the Association
  for Computational Linguistics (Volume 2: Short Papers)}, pages 110--117,
  Vancouver, Canada. Association for Computational Linguistics.

\bibitem[{Rasooli and Tetreault(2015)}]{rasooli2015yara}
Mohammad~Sadegh Rasooli and Joel Tetreault. 2015.
\newblock Yara parser: A fast and accurate dependency parser.
\newblock \emph{arXiv preprint arXiv:1503.06733}.

\bibitem[{Shi et~al.(2017)Shi, Huang, and Lee}]{D17-1002}
Tianze Shi, Liang Huang, and Lillian Lee. 2017.
\newblock Fast(er) exact decoding and global training for transition-based
  dependency parsing via a minimal feature set.
\newblock In \emph{Proceedings of the 2017 Conference on Empirical Methods in
  Natural Language Processing}, pages 12--23. Association for Computational
  Linguistics.

\bibitem[{Straka et~al.(2015)Straka, Hajic, Strakov{\'a}, and
  Hajic~jr}]{straka2015parsing}
Milan Straka, Jan Hajic, Jana Strakov{\'a}, and Jan Hajic~jr. 2015.
\newblock Parsing universal dependency treebanks using neural networks and
  search-based oracle.
\newblock In \emph{International Workshop on Treebanks and Linguistic Theories
  (TLT14)}, pages 208--220.

\bibitem[{Titov and Henderson(2007)}]{Titov07Conll}
Ivan Titov and James Henderson. 2007.
\newblock Fast and robust multilingual dependency parsing with a generative
  latent variable model.
\newblock In \emph{Proc.\ of the CoNLL shared task. Joint Conf. on Empirical
  Methods in Natural Language Processing and Computational Natural Language
  Learning (EMNLP-CoNLL)}, Prague, Czech Republic.

\bibitem[{Vacariu(2017)}]{vacariu2017high}
Andrei~Vlad Vacariu. 2017.
\newblock \emph{A high-throughput dependency parser}.
\newblock Ph.D. thesis, Applied Sciences: School of Computing Science, Simon
  Fraser University.

\bibitem[{Vilares and G{\'o}mez-Rodr{\'\i}guez(2017)}]{vilares2017non}
David Vilares and Carlos G{\'o}mez-Rodr{\'\i}guez. 2017.
\newblock A non-projective greedy dependency parser with bidirectional {LSTM}s.
\newblock \emph{Proceedings of the CoNLL 2017 Shared Task: Multilingual Parsing
  from Raw Text to Universal Dependencies}, pages 152--162.

\bibitem[{Volokh(2013)}]{volokh2013performance}
Alexander Volokh. 2013.
\newblock \emph{Performance-Oriented Dependency Parsing}.
\newblock Doctoral dissertation, Saarland University, Saarbr{\"u}cken, Germany.

\bibitem[{Xue et~al.(2005)Xue, Xia, Chiou, and Palmer}]{xue2005penn}
Naiwen Xue, Fei Xia, Fu-Dong Chiou, and Marta Palmer. 2005.
\newblock The {P}enn {C}hinese {T}reebank: {P}hrase structure annotation of a
  large corpus.
\newblock \emph{Natural language engineering}, 11(2):207--238.

\bibitem[{Zhang and Nivre(2011)}]{zhang2011transition}
Yue Zhang and Joakim Nivre. 2011.
\newblock Transition-based dependency parsing with rich non-local features.
\newblock In \emph{Proceedings of the 49th Annual Meeting of the Association
  for Computational Linguistics: Human Language Technologies: short
  papers-Volume 2}, pages 188--193. Association for Computational Linguistics.

\end{thebibliography}

\end{document}